\documentclass[sigconf,nonacm]{acmart}
\usepackage{algorithm}
\usepackage{algpseudocode}
\usepackage{url}
\usepackage{float}
\usepackage{stfloats}
\usepackage{makecell}
\usepackage{textcomp}

\AtBeginDocument{%
  }

\setcopyright{none}
\acmDOI{}
\acmISBN{}
\acmConference[]{}{}{}
\acmBooktitle{}

\begin{document}

\title{Fast Linear Solvers via AI-Tuned Markov Chain Monte Carlo-based Matrix Inversion}

\author{Anton Lebedev}
\authornote{Both authors contributed equally to this research.}
\email{anton.lebedev@stfc.ac.uk}
\author{Won Kyung Lee}
\authornotemark[1]
\email{wonkyung.lee@stfc.ac.uk}
\affiliation{%
  \institution{STFC Hartree Centre}
  \city{Warrington}
  \country{United Kingdom}
}

\author{Soumyadip Ghosh}
\affiliation{%
  \institution{IBM Research}
  \city{Yorktown Heights}
  \state{New York}
  \country{USA}}
\email{ghoshs@us.ibm.com}

\author{Olha I. Yaman}
\affiliation{%
  \institution{STFC Hartree Centre}
  \city{Warrington}
  \country{United Kingdom}
}
\email{olha.ivanyshyn-yaman@stfc.ac.uk}

\author{Vassilis Kalantzis}
\affiliation{%
 \institution{IBM Research}
 \city{Yorktown Heights}
 \state{New York}
 \country{USA}}
\email{vkal@ibm.com}

\author{Yingdong Lu}
\affiliation{%
  \institution{IBM Research}
  \city{Yorktown Heights}
  \state{New York}
  \country{USA}}
\email{yingdong@us.ibm.com}

\author{Tomasz Nowicki}
\affiliation{%
  \institution{IBM Research}
  \city{Yorktown Heights}
  \state{New York}
  \country{USA}}
\email{tnowicki@us.ibm.com}

\author{Shashanka Ubaru}
\affiliation{%
  \institution{IBM Research}
  \city{Yorktown Heights}
  \state{New York}
  \country{USA}}
\email{shashanka.ubaru@ibm.com}

\author{Lior Horesh}
\affiliation{%
  \institution{IBM Research}
  \city{Yorktown Heights}
  \state{New York}
  \country{USA}}
\email{lhoresh@us.ibm.com}

\author{Vassil Alexandrov}
\affiliation{%
  \institution{STFC Hartree Centre}
  \city{Warrington}
  \country{United Kingdom}}
\email{vassil.alexandrov@stfc.ac.uk}

\renewcommand{\shortauthors}{Lebedev et al.}

\begin{abstract}
Large, sparse linear systems are pervasive in modern science and engineering, and Krylov subspace solvers are an established means of solving them. Yet convergence can be slow for ill-conditioned matrices, so practical deployments usually require preconditioners. Markov chain Monte Carlo (MCMC)-based matrix inversion can generate such preconditioners and accelerate Krylov iterations, but its effectiveness depends on parameters whose optima vary across matrices; manual or grid search is costly. We present an AI-driven framework recommending MCMC parameters for a given linear system. A graph neural surrogate predicts preconditioning speed from $A$ and MCMC parameters. A Bayesian acquisition function then chooses the parameter sets most likely to minimise iterations. On a previously unseen ill-conditioned system, the framework achieves better preconditioning with 50\% of the search budget of conventional methods, yielding about a 10\% reduction in iterations to convergence. These results suggest a route for incorporating MCMC-based preconditioners into large-scale systems.
\end{abstract}



\received[accepted]{5 September 2025}

\maketitle
\noindent\small
\textbf{Author's version.} \textcopyright 2025 The Authors. Publication rights licensed to ACM. One or more authors of this work are employees or contractors of a national government. This is the author's version of the work. It is posted here for your personal use; not for redistribution. The definitive Version of Record was published in \textit{SC Workshops '25, November 16--21, 2025, St Louis, MO, USA}, 
\href{https://doi.org/10.1145/3731599.3767543}{https://doi.org/10.1145/3731599.3767543}.
\normalsize

\section{Introduction}
Large and sparse systems of linear equations arise daily in computational fluid dynamics, structural analysis, plasma physics and countless other branches of modern science and engineering. Krylov subspace methods such as Conjugate Gradient (CG), Generalized Minimal RESidual method (GMRES) method, and their variants remain the work-horses for solving such systems because they require only matrix–vector products and a modest memory footprint \cite{saad2003iterative}. Nevertheless, their convergence can be quite slow when the matrix is ill-conditioned, which is the case when the matrix represents a differential operator discretised on a fine mesh. In such cases, practical applications rely on preconditioners that improve the spectral properties of the system.

Classical algebraic preconditioners such as Incomplete LU (ILU) and Incomplete Cholesky (IC) factorisations are powerful yet may fail or become prohibitively expensive for highly irregular matrices. In response, Markov-chain Monte-Carlo (MCMC) matrix-inversion (MI) techniques have resurfaced as an appealing alternative preserving the sparsity of the matrix while offering a high degree of embarrassing parallelism. Despite these merits, the performance of MCMC preconditioning is acutely sensitive to the choice of hyperparameters, and optimal values generally differ from one linear system to the next. Exhaustive or grid searches over the parameter space require many expensive solver runs, thus undermining the practical advantage that MCMC was meant to provide.

In this paper, we tackle the bottleneck of algorithmic parameter tuning for MCMC preconditioners. In particular, we propose an AI-assisted framework that intelligently selects MCMC parameters for any given sparse system. At its core, our framework leverages a graph neural surrogate model. Given an iteration matrix $A$ and a set of candidate MCMC parameters, our model predicts the expected reduction in the number of iterations to achieve convergence compared to the unpreconditioned case. A Bayesian acquisition function then exploits the surrogate’s statistical nature, in each batch, those parameter sets which are most likely to minimise the expected iteration count. Applied to previously unseen matrices, the framework exceeds baseline preconditioning performance with only 50\% of the search budget demanded by conventional methods and delivers about 10\% fewer number of steps to convergence. These results indicate a practical route towards deploying MCMC preconditioners in large-scale linear-solver pipelines with minimal manual intervention.

The remainder of the paper is organised as follows. Section 2 reviews related work, covering both advances in preconditioning and MCMC-based MI. Section 3 introduces our proposed framework, including details of the surrogate model architecture and Bayesian selection loop. Section 4 details the experimental set up, while Section 5 discusses results and limitations. Section 6 concludes with avenues for future research.

\section{Literature Review}
Early work on algebraic preconditioning concentrated on deterministic approximations of $A^{-1}$. Incomplete factorisations \cite{saad1994ilut} such as ILU and IC remain a staple of various large-scale scientific and engineering simulations. However, they are difficult to pipeline on modern hardware while ILU may break down for indefinite matrices \cite{dillon2018hierarchical,xu2022pargemslr}. Sparse Approximate Inverse (SPAI) schemes address this parallelism bottleneck \cite{grote1997parallel}, constructing an explicit sparse stand-in for $A^{-1}$ that can be applied via Sparse Matrix-Vector multiplications (SpMV)—an operation that parallelises well. Stochastic methods based on MCMC sidestep the triangular-solve bottleneck of factorisation-type preconditioners by estimating columns of $A^{-1}$ through independent random walks \cite{strassburg2013monte}. Because each walk evolves independently, the work decomposes into parallel tasks that can scale almost linearly on modern accelerator hardware \cite{lebedev2018advanced}. The approach has proved viable on large real-world test cases, including electromagnetic, climate, plasma physics simulations \cite{sahin2021usability}, while ongoing algorithmic advances—most recently the regenerative formulation that collapses multiple hyperparameters into a single transition budget parameter—continue to improve robustness and variance control \cite{ghosh2024regenerative}. Despite their architectural advantages, these methods have a critical practical flaw. Their effectiveness hinges on a set of hyperparameters whose optimal values are strongly matrix-dependent. Finding them often requires costly, manual trial-and-error, creating a significant tuning bottleneck.

On the other hand, there have been recent attempts to develop learning-based matrix preconditioners that aim to infer a good approximation of the inverse from data. In particular, \cite{chen2025graph} suggested a graph neural preconditioner that generates a preconditioner for a matrix in a linear system through a black box graph neural network model. Another significant area of interest is the specialisation of learning-based preconditioners for systems derived from Partial Differential Equations (PDEs), typically by utilising additional geometric or physical information from the underlying physical problem as additional machine learning model inputs. For instance, some methods learn from the problem's geometry by using the physical grid coordinates as model inputs \cite{trifonov2024learning, luo2024neural}, whilst others learn from its physical by directly incorporating the PDE's coefficients \cite{li2023learning}. Such PDE-tailored methods are further constrained by their application-specific design. More fundamentally, learning-based matrix preconditioners suffer from inherit limitations related to their "black-box" nature, such as the difficulty in diagnosing and remedying poor performance on a specific matrix. This hinders their practical adoption, especially for mission-critical applications where reliability, stability and explainability are of paramount importance.

\section{Methodology}

In this paper, we consider the linear system $Ax=b$, where $A\!\in\!\mathcal A\subset \mathbb{R}^{n\times n}$. Our aim is to obtain a fast approximation of the inverse $A^{-1}$ so that the system can be solved more efficiently. As a representative example, we adopt the MCMC-based MI schemes of the prior literature \cite{lebedev2018advanced, sahin2021usability}; more recent variants such as \cite{ghosh2024regenerative} could be also employed. An MCMC-based MI method requires a vector of algorithmic parameters $x_{\text{M}} \in \mathcal{X}_M$ and returns a preconditioner $P\approx A^{-1}$. 
We then solve $PAx = Pb$, where the Krylov solver will typically converge faster due to the lower condition number of $PA$. 
 
Here we aim to identify parameters $x_{\text{M}}$ of the MCMCMI method that will minimise the overall time-to-solution of the system, the time required to create the preconditioner and solve the system, accounting for matrix dependence via matrix $A$ and its features $x_A\in\mathcal{X}_A$. 
To this end, we construct a surrogate model that, given $(A, x_A, x_{\text{M}})$, predicts the resulting expected preconditioning speed $\mu(A, x_{\text{M}})$. Specifically, the surrogate model outputs the predicted mean $\widehat{\mu}$ together with an uncertainty estimate $\widehat{\sigma}$. The optimal MCMC parameters are then defined by \(x^{*}_{\text{M}}(A)=\arg\min_{x_{\text{M}}}\mu(A,x_{\text{M}})\), and are selected via an acquisition function that balances exploration of the parameter space with exploitation of the surrogate's current best estimate.

\subsection{Graph Neural Surrogate Model}
We employ a graph neural network \(f_{\boldsymbol{\theta}}\) as the surrogate model because its message passing operations are size-invariant so that the model can take varying sizes of matrices as parts of inputs. We design the graph neural surrogate model that extracts information directly from the matrix $A$ and augments it with inexpensive matrix features $x_A$ and the candidate MCMC parameter vector $x_{\text{M}}$. For this, we construct a weighted and directed graph \(G=(V,x_{V},E,w_{E})\) from the matrix \(A\in\mathbb R^{n\times n}\), whose vertex set \(V=\{1,\dots,n\}\) represents the rows of \(A\). An edge \((i,j)\in E\) exists iff \(A_{ij}\neq0\) and carries weight \(w_{E}(i,j)=A_{ij}\). Each vertex stores the unweighted row degree \(x_{V}(i)=\deg(i)=\lvert\{\,j:A_{ij}\neq0\,\}\rvert\).

In addition to graph data, matrix dependence is also accounted for via matrix features $x_A \in \mathcal{X}_A$ that are cheap to compute, such as the norms, sparsity and symmetricity. 
All features are standardised—each value is rescaled to zero mean and unit variance—so that they contribute on a comparable scale during training.

The graph neural surrogate model receives the triplet \((G,x_{A},x_{\text{M}})\) and processes each component separately before fusion. A stack of \(l_{g}\) message passing layers extracts a graph embedding \(h_{g}\) from \(G\) \cite{battaglia2018relational}. Although numerous message-passing formulations exist, most of them follow the similar basic pattern: at each layer, every node aggregates information from its neighbours and itself, summarises (i.e., pools) these messages, and then applies a non-linear transformation to produce an updated representation. Several modern graph neural layer architectures are explored, and the most suitable one is selected through extensive experimentation. In parallel, \(l_{A}\) and \(l_{M}\) fully connected (FC) layers transform \(x_{A}\) and \(x_{\text{M}}\) into embeddings \(h_{A}\) and \(h_{M}\). Layer normalisation is applied in both the message passing layers and FC stacks to stabilise training and mitigate covariate shift, while ReLU provides the non-linear activation. The three latent representations \(h_{g}\), \(h_{A}\), and \(h_{M}\) are then concatenated and passed through \(l_{c}\) FC layers with dropout, producing a vector \(h_{\text{combined}}\). Finally, two linear heads provide the predicted mean and standard deviation of the MCMC preconditioning performance metric:
\begin{equation}\label{mu_sigma_MCMC_performance}
    \widehat{\mu}\;=\;\operatorname*{ReLU}\!\bigl(W_{\mu}h_{\text{combined}} + b_{\mu}\bigr),
    \widehat{\sigma}\;=\;\ln\!\bigl(1 + e^{\,W_{\sigma}h_{\text{combined}} + b_{\sigma}}\bigr),
\end{equation}
where \(W_{\mu},\,b_{\mu},\,W_{\sigma}\) and \(b_{\sigma}\) are trainable weights and biases. The former expression applies a ReLU to obtain the predicted mean, while the latter employs the soft-plus transform \(\ln(1+e^{z})\) to ensure a strictly positive standard deviation. We model the MCMC preconditioning performance as a Gaussian distribution with mean \(\widehat{\mathcal\mu}\) and variance \(\widehat{\mathcal\sigma}^{2}\).

Given a training dataset \(\mathcal{D}=\{(G_{i},x_{A,i},x_{\text{M},i},\bar{y}_{i},s_{i})\}_{i=1}^{N}\), where \(\bar{y}_{i}\) and \(s_{i}\) are the sample mean and sample standard deviation of the repeated solver runs for the \(i\)-th input, we learn the surrogate model parameters by minimising a Mean Squared Error objective
\begin{equation}\label{MSE_objective}
    \mathcal{L}(\boldsymbol{\theta})
    =\frac{1}{N}\sum_{i=1}^{N}
    \bigl[(\widehat{\mu}_{i}-\bar{y}_{i})^{2} + (\widehat{\sigma}_{i}-s_{i})^{2}\bigr].
\end{equation}
A Gaussian negative log-likelihood could be also considered as an alternative, but very small \(s_{i}\) values could make that objective numerically unstable.

\subsection{Acquisition Function}
In most cases, we never get enough evaluations to exhaustively scan the MCMC parameter space. We therefore rely on Bayesian Optimisation (BO) to determine the next parameter vectors \(x_{\text{M}}\) to test. The surrogate model \(f_{\boldsymbol{\theta}}\) provides, for each candidate, a predictive mean \(\widehat{\mu}\) and a predictive uncertainty \(\widehat{\sigma}\) as in \ref{mu_sigma_MCMC_performance}. An acquisition function balances two objectives: it exploits regions in which the predicted mean \(\widehat{\mu}\) is already low, while it simultaneously explores regions with high predictive uncertainty \(\widehat{\sigma}\), as those areas may still conceal better solutions. Past observations on related matrices, for example, smaller matrices representing the same differential operator, allow the surrogate model to transfer knowledge, thereby reducing uncertainty for similar systems.

While a variety of acquisition functions are available, we adopt Expected Improvement (EI) because it has been shown to deliver consistently lower simple regret than others, for instance, confidence-bound methods \cite{srinivas2009gaussian} across a large suite of benchmarks \cite{merrill2021empirical}. EI naturally balances exploitation and exploration while relying on a single and intuitive exploration parameter \(\xi\) \cite{mockus1994application, mockus1998application}. Setting \(\xi=0\) yields pure exploitation, whereas values in the range \(0.01\!-\!0.10\) gradually favour uncertain regions of the search space. Moreover, for a Gaussian surrogate posterior EI has the closed form
\begin{equation}\label{EI_function}
    \operatorname{EI}(x_{\text{M}})
    =\bigl(y_{\min}-\widehat{\mu}-\xi\bigr)\,
      \Phi\!\Bigl(\tfrac{y_{\min}-\widehat{\mu}-\xi}{\widehat{\sigma}}\Bigr)
    +\widehat{\sigma}\,
      \varphi\!\Bigl(\tfrac{y_{\min}-\widehat{\mu}-\xi}{\widehat{\sigma}}\Bigr),
\end{equation}
where \(y_{\min}\) is the best MCMC preconditioning performance metric observed so far and \(\Phi\) and \(\varphi\) are the standard normal Cumulative Distribution Function and Probability Density Function, respectively. The first term measures the expected drop below the current best exploitation, whereas the second term rewards large predictive variance exploration. Because EI is differentiable with respect \(x_{\text{M}}\), we can maximise it efficiently with first-order optimisers. In practice, we minimise the negative EI using the gradient-based quasi-Newton method L-BFGS-B \cite{byrd1995limited}. At every step the candidate \(x_{\text{M}}\) is fed through the surrogate model; back-propagation supplies the exact gradient $\nabla_{x_{\text{M}}}\left[-\operatorname{EI}(x_{\text{M}})\right]$, which L-BFGS-B exploits to build curvature information and update the iterate. Algorithm \ref{alg:bo_loop} summarises the complete optimisation loop.

\begin{algorithm}[t]
\caption{Bayesian tuning loop for MCMC parameter selection}
\label{alg:bo_loop}
\small
\begin{algorithmic}
\Require evaluated matrix set $\mathcal A_{\text{train}}$, total budget $B$, batch size $k$
\State Initialise $\mathcal D_{0}$ with coarse grid-search records $(A,x_{\text{M}},\bar y,s)$
\For{$t=0,1,\dots$ \textbf{until} $\lvert\mathcal D_t\rvert = B$}
    \State Fit surrogate $f_{\boldsymbol{\theta}}$ on $\mathcal D_t$
    \ForAll{$A \in \mathcal A_{\text{train}}$}
        \For{$j=1$ \textbf{to} $k$}
            \State draw initial $x_{\text{M}}^{(j,\text{init})}$
            \State $x_{\text{M}}^{(j)}\!\gets$ \textbf{L--BFGS--B} maximise EI$(x_{\text{M}};A)$ starting from $x_{\text{M}}^{(j,\text{init})}$
            \State Run MCMC $+$ Krylov solver (e.g., GMRES) with $x_{\text{M}}^{(j)}$
            \State Record $(A,x_{\text{M}}^{(j)},\bar y,s)$ and append to $\mathcal D_{t+1}$
        \EndFor
    \EndFor
\EndFor
\State \Return $x_{\text{M}}^{\star}(A) \;=\;\displaystyle\arg\max_{x_{\text{M}}}\mathrm{EI}(x_{\text{M}};A)$ given $A\!\in\!\mathcal A$
\end{algorithmic}
\end{algorithm}

\section{Experiments}
\subsection{MCMC Preconditioning}
To benchmark our tuning framework we adopt the advanced MCMC-based MI preconditioner of \cite{lebedev2018advanced, sahin2021usability}. The method is governed by three continuous algorithmic parameters \(x_{\text{M}}=(\alpha,\varepsilon,\delta)\):
\begin{itemize}
\item \(\alpha\in\mathbb R_{>0}\): a matrix perturbation parameter to scale the added diagonal of \(A\) so that the Neumann-series preconditioner converges;
\item \(\varepsilon\in(0,1]\): a stochastic error that determines the maximum number of independent Markov chains;
\item \(\delta\in(0,1]\): a truncation error that determines the maximum walk length of a Markov chain.
\end{itemize}
In addition, $x_{\text{M}}$ includes a categorical variable for the Krylov solver type. Due to the limited size of our dataset, we do not attempt to provide a recommendation of the solver type here.
Two matrix-independent settings are fixed: the filling factor, which controls the number of non-zeros retained in the preconditioner, and the truncation threshold. The preconditioner's filling factor is set to $2\phi(A)$, where $\phi(A)$ is that of the original matrix $A$, whereas the truncation threshold is set arbitrarily to $10^{-9}$ to avoid introducing truncation of the preconditioners. Also, all preconditioners are obtained using a hybrid MPI+OpenMP code running on a single node utilising 2 MPI processes with 4 threads per process. The preconditioned system is then solved with GMRES or BiCGStab; when the matrix $A$ is symmetric positive definite, we also employ CG. Also, we define the MCMC preconditioning performance metric 
\begin{equation}\label{performance_metric}
    \text{y}(A,x_{\text{M}})=
    \frac{\#\,\text{of steps with preconditioner}}
         {\#\,\text{of steps without preconditioner}},
\end{equation}
so that the optimiser seeks the \(x_{\text{M}}\) minimising this ratio for every matrix.

\subsection{Dataset}
We built the dataset from 11 sparse matrices summarised in Table \ref{tab:matrices}.  
Each entry appears with its dimension~$n$, symmetricity, and condition number $\kappa(A)\!=\!\lVert A\rVert_{2}\lVert A^{-1}\rVert_{2}$. The selection covers the archetypal 2D Finite-Difference (FD) Laplacian matrix as a representative of FD methods and symmetric positive-definite matrices, \verb|a0XXXX| matrices, representing asymmetric differential operators from plasma physics discretised using finite elements at various mesh resolutions, a finite-element discretisation of an unsteady advection-diffusion problem with varying mesh resolutions, \texttt{PDD\_RealSparse}, and a representative of systems occurring in climate simulations (\texttt{nonsym\_r3\_a11}). In FD or finite-element discretisations on shape-regular meshes, the condition number of the matrix for an $m$-th order PDE operator scales with the mesh width $h$ as $O\left(h^{-m}\right)$. In particular, $O\left(h^{-2}\right)$ scaling is illustrated in Table \ref{tab:matrices} for the 2D FD Laplacian matrix. A large condition number can severely degrade the performance of iterative solvers, making effective preconditioning essential for maintaining computational efficiency.

To obtain the basis dataset for training we used a $4\times4\times4$ grid of parameters
\(\alpha\!\in\!\{1,2,4,5\}\),
\(\varepsilon\!\in\!\{1/2,1/4,1/8,1/16\}\),
\(\delta\!\in\!\{1/2,1/4,1/8,1/16\}\) with
each \(\alpha,\varepsilon,\delta\) configuration executed ten times with GMRES and BiCGStab. The resulting sample mean and standard deviation of the performance metric 
y$(A,x_{\text{M}})$ constitute one labelled datum per solver. Hence every matrix contributed 64 samples per solver (128 in total for the two-solver case). 
The symmetric Laplace matrices were additionally run with CG at
\(\alpha=0.1\).
A few samples with near-zero~$\alpha$ were added to expose the surrogate to divergence scenarios. In total, the dataset for training and validation contains 1,318 labelled points, which were split 80\%/20\% into training and validation sets.

On the other hand, generalisability was assessed on the higher-order \texttt{unsteady\_adv\_diff\_order2\_0001}
(\(\kappa\simeq6.6\times10^{6}\)), a substantially harder system than its order-1 counterpart in the training phase; success here demonstrates that information transfers to an unseen ill-conditioned system.

\begin{table}[t]
\centering\small
\setlength{\tabcolsep}{3pt}
\caption{Matrix set used for this study}
\label{tab:matrices}
\begin{tabular}{@{}lcccc@{}}
\toprule
Matrix & Dimension & \makecell{Symmet- \\ ricity} & $\kappa(A)$ & $\phi(A)$ \\ \midrule
2DFDLaplace\_16 & 225   & Yes & $1.0\times10^{2}$ & 0.042\\
2DFDLaplace\_32 & 961   & Yes & $4.1\times10^{2}$ & 0.001\\
2DFDLaplace\_64 & 3,969 & Yes & $1.7\times10^{3}$ & 0.0024\\
2DFDLaplace\_128& 16,129& Yes & $6.6\times10^{3}$ & 0.0006\\[2pt]
nonsym\_r3\_a11 & 20,930& No  & $1.9\times10^{4}$ & 0.0044\\
a00512          & 512   & No  & $1.9\times10^{3}$ & 0.059\\
a08192          & 8,192 & No  & $3.2\times10^{5}$ & 0.0007\\
unsteady\_adv\_diff\_order1\_0001& 225 & No & $4.1\times10^{6}$ & 0.646\\
unsteady\_adv\_diff\_order2\_0001& 225 & No & $6.6\times10^{6}$ & 0.646\\[2pt]
PDD\_RealSparse\_N64& 64  & No & $1.3\times10^{1}$ & 0.1\\
PDD\_RealSparse\_N128& 128 & No & $5.0$ & 0.1\\
PDD\_RealSparse\_N256& 256 & No & $7.0$ & 0.1\\ \bottomrule
\end{tabular}
\end{table}

\subsection{Hyperparameters}
We performed extensive hyperparameter tuning for both the graph neural surrogate model and the acquisition function to maximise predictive accuracy and optimisation efficacy. For the graph convolutional architecture, we considered six representative message-passing mechanisms: GATv2\cite{brody2022how}, Graph Transformer\cite{shi2021masked}, GMMConv \cite{monti2017geometric}, EdgeConv\cite{wang2019dynamic}, GINE\cite{Hu*2020Strategies}, and PNA\cite{corso2020principal}. These were combined with three neighbourhood aggregation strategies: MultiAggregation \cite{corso2020principal}, MeanAggregation, and DeepSetsAggregation\cite{buterez2022graph}. We searched over hidden dimensions \{32, 64, 128, 256, 512\} and up to four message-passing layers. For the auxiliary inputs $x_A$ and $x_{\text{M}}$, FC layers were employed with one to four layers, and hidden dimensions chosen from \{8, 16, 32, 64\} for $x_A$ and \{4, 8, 16, 32\} for $x_{\text{M}}$. The concatenated embedding was passed through another FC block with hidden dimensions in \{32, 64, 128, 256, 512\} and up to four layers. We fixed the batch size at 128.

Continuous hyperparameters included the learning rate, sampled from a log-uniform distribution between $10^{-4}$ and $10^{-1}$, the weight decay parameter from $10^{-6}$ to $10^{-3}$, and dropout rates uniformly from 0 to 0.2. Hyperparameter optimisation was performed using the Tree-structured Parzen Estimator \cite{bergstra2011algorithms}. We used the Asynchronous Successive Halving Algorithm scheduler \cite{li2020system} for early stopping and resource-efficient scheduling, with a maximum of 150 epochs, a grace period of 20, and a reduction factor of 3. A total of 30 trials were launched, each corresponding to a different model configuration. In the acquisition function, we tested both a balanced strategy with $\xi=0.05$ and an exploration-heavy strategy with $\xi=1.0$.

\subsection{Experimental Results}
We assess the proposed pipeline on an unseen, highly ill-conditioned testing matrix. The graph neural surrogate was trained by a single NVIDIA V100 GPU (32 GB), and all other experiments were executed on CPUs.

The surrogate trained on the training dataset is referred to as \textsc{Pre-BO Model}. The best graph neural surrogate model was selected through the Bayesian Optimisation-based hyperparameter optimisation (HPO), based on the average performance across three random seeds. The selected surrogate model architecture include learning rate as $1.848 \times 10^{-3}$, weight decay parameter as 1, a single Edge Convolutional layer \cite{wang2019dynamic} and mean neighbourhood aggregation function with 256 hidden dimension for graph embedding, while a single FC layer with 64 dimension for embedding of the matrix feature \(x_{A}\) and three FC layers with 16 hidden dimension for MCMC algorithmic parameter \(x_{\text{M}}\) embedding. Also, two FC layers were added to represent the combined representation with 128 hidden dimension. The Adam optimiser \cite{kingma2015adam} was used for model training. Although model training and model selection with HPO required approximately seven hours of compute time, such cost is expected to be amortised when applying the framework to large-scale matrices, where reduced solution cost can yield substantial overall savings.

To assess how the GNN surrogate evolves when augmented with new, targeted data, we performed one round of BO using the EI acquisition function (\ref{EI_function}) with two settings: $\xi=0.05$ (balanced search) and $\xi=1.00$ (exploration search).  
The \textsc{Pre-BO Model} was used to recommend a batch of 32 candidate $x_{\text{M}}$ vectors for each BO strategy, for which MCMC preconditioning metrics were measured (10 replicates each). These new measurements were combined with the original training dataset to retrain the surrogate, producing the \textsc{BO-enhanced Model}. During retraining, we reused the hyperparameters selected for the \textsc{Pre-BO Model} and re-optimised only the model weights. Model retraining required approximately an hour of compute time. Performance was evaluated on an unseen, ill-conditioned testing matrix \path{unsteady_adv_diff_order2_0001}, using experimental results from a grid search over 64 distinct $x_{\text{M}}$ vectors (10 replications each, 640 observations in total).

First, we assessed the reliability of the surrogate models' uncertainty estimates, whether the predicted confidence intervals capture the observed variability at the expected rate. This calibration assessment reveals whether the model is overconfident (intervals too narrow) or underconfident (intervals too wide), which is critical when the surrogate model is used to guide BO decisions.

\begin{figure}[htbp]
    \centering
    \includegraphics[width=\columnwidth]{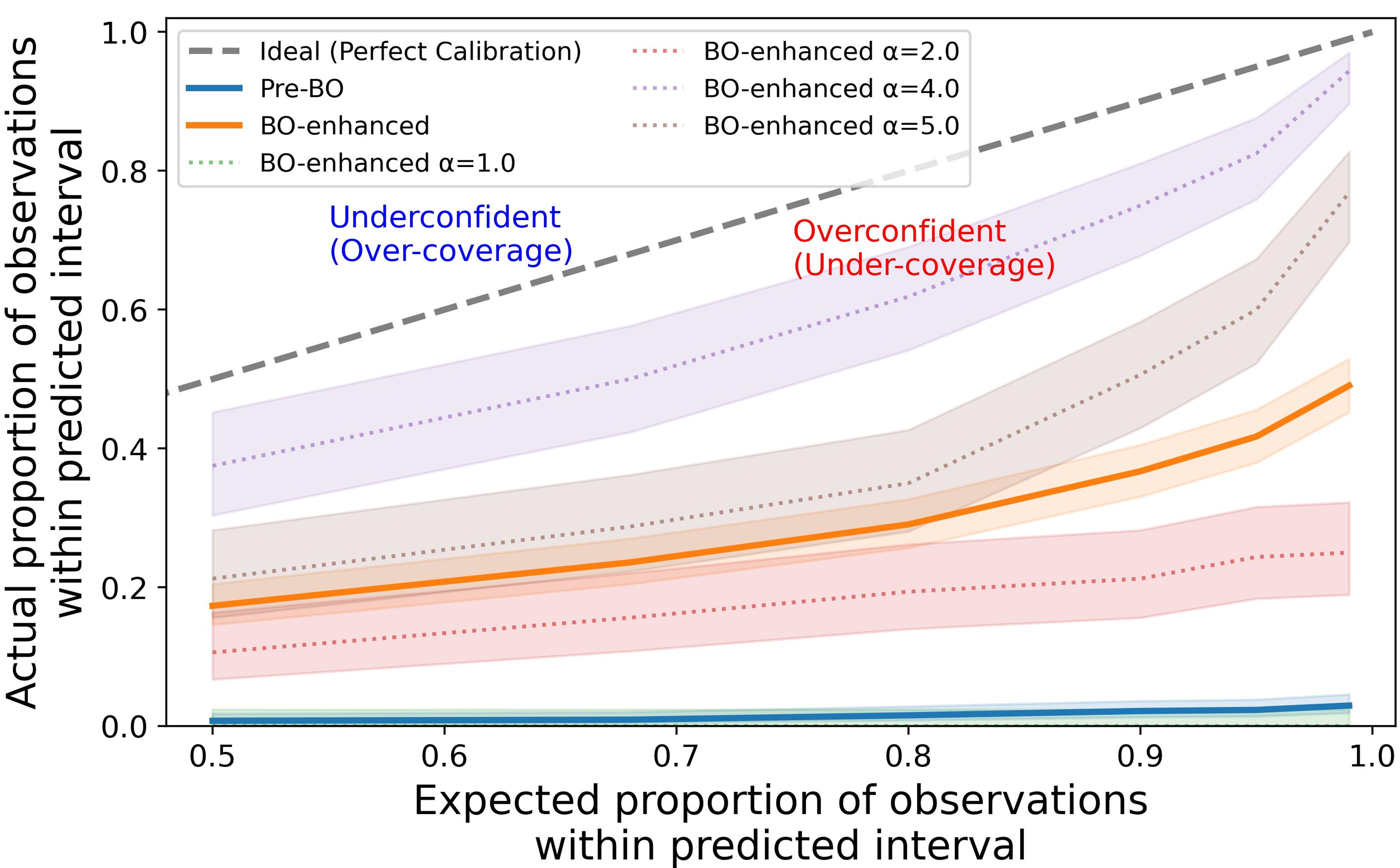}
    \caption{Calibration plot comparing predicted and observed coverage probabilities across multiple confidence levels}
    \Description{Calibration plot comparing predicted and observed coverage probabilities across multiple confidence levels}
    \label{fig:calibration_PI_plot}
\end{figure}

Figure~\ref{fig:calibration_PI_plot} shows calibration curves comparing the expected proportion of observations within the predicted interval (x-axis) to the actual proportion observed (y-axis) for the corresponding confidence levels $\tau \in \{0.50, 0.68, 0.80, 0.90, 0.95, 0.99\}$. For each $\tau$, the symmetric prediction interval was defined as
\begin{equation}\label{prediction_interval}
    \left[\hat\mu_j - z_{(1+\tau)/2}\,\hat\sigma_j,\; \hat\mu_j + z_{(1+\tau)/2}\,\hat\sigma_j\right],
\end{equation}
where $(\hat\mu_j,\hat\sigma_j)$ are the surrogate model's predicted mean and standard deviation, and $j$ indexes the $640$ individual observations ($64$ distinct $x_{\text{M}}$, each with $10$ replicates), with $(\hat\mu_j, \hat\sigma_j)$ identical within replicates of the same $x_{\text{M}}$. The empirical coverage $\hat p$ was computed as the proportion of observations $y_j$ falling within this interval.

To quantify the sampling uncertainty in $\hat p$, we computed the two-sided Wilson score 95\% confidence interval for a binomial proportion \cite{wilson1927probable}:
\begin{equation}\label{Wilson_interval}
    \mathrm{CI}_{\text{Wilson}}(\hat p) \;=\;
    \frac{\hat p + \frac{z^2}{2n} \pm z\sqrt{\frac{\hat p(1-\hat p)}{n} + \frac{z^2}{4n^2}}}
    {\,1+\frac{z^2}{n}\,},
\end{equation}
where $n$ is the number of observations and $z=z_{0.975}$. This method is preferred over the normal approximation because it produces well-behaved bounds in $[0,1]$, even for small $n$ or extreme proportions. Shaded bands in Figure~\ref{fig:calibration_PI_plot} represent these Wilson intervals. According to the plot, the \textsc{Pre-BO Model} exhibits clear under-coverage (overconfidence), with curves lying below the ideal diagonal. After a single BO round, the \textsc{BO-enhanced Model} shifts markedly closer to the diagonal, indicating improved calibration. In particular, for higher $\alpha$ values ($\alpha=4.0$ and $\alpha=5.0$), coverage approaches the ideal line, and the improvement is statistically significant according to the Wilson intervals (\ref{Wilson_interval}). On the other hand, 
when the iteration matrix is ill-conditioned, lower values of $\alpha$ generally result to similar measurements, which in turn limits the learning ability of the surrogate model.

\begin{figure*}[htbp]
    \centering
    \includegraphics[width=0.8\textwidth]{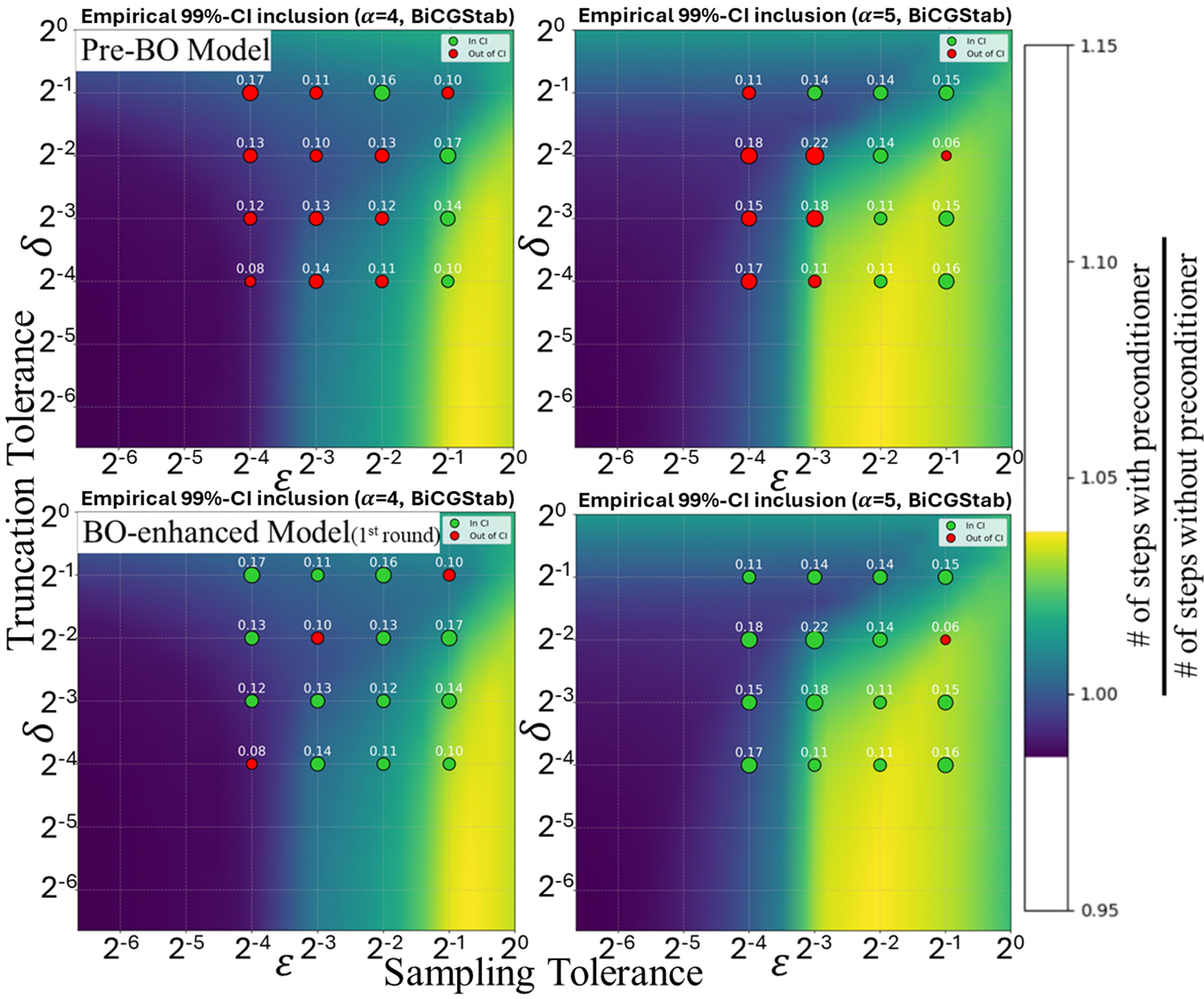}
    \caption{Confidence interval inclusion results showing the improved prediction accuracy of the surrogate model after BO retraining (BO-enhanced, bottom) compared to the baseline (Pre-BO, top)}
    \Description{Confidence interval inclusion results showing the improved prediction accuracy of the surrogate model after BO retraining (BO-enhanced, bottom) compared to the baseline (Pre-BO, top)}
    \label{fig:accuracy_CI}
\end{figure*}

Further we examined whether the surrogate model's predicted mean lies within the empirical confidence interval for each set of MCMC algorithmic parameters $x_{\text{M}}$. Specifically, for each of the 64 distinct $x_{\text{M}}$, we computed the sample mean $\bar{y}$, standard deviation $s$ and 99\% confidence interval of the preconditioning performance metric $y(A,x_{\text{M}})$ across 10 replications. We then investigated whether the model's predicted mean falls inside this empirical interval. This pointwise coverage analysis focuses on the accuracy of the predicted central value at each algorithmic parameter point, rather than on the coverage of the observed data by the model's own predicted intervals. According to the plot shown in Figure~\ref{fig:accuracy_CI}, the \textsc{Pre-BO Model} (top row) does not represent reality well, as its mean frequently lies outside of the confidence interval. In contrast, the \textsc{BO-enhanced model} (bottom row) achieves substantially higher inclusion over broad regions across the $(\varepsilon,\delta)$ grid for higher alphas $\alpha\!\in\!\{4.0,5.0\}$. 
The heatmaps show, that contrary to prior assumptions \cite{lebedev2018advanced}, $\varepsilon$ and $\delta$ do not contribute symmetrically to the success of the preconditioner. We observe that, given a truncation error $\delta$, a successful preconditioner is obtained if $\varepsilon\lessapprox \delta$, with this condition being more pronounced at larger perturbations $\alpha$. Conversely, for a fixed stochastic error $\varepsilon$, thus a fixed number of Markov chains, a larger $\delta$ and thus shorter chains are preferable. Since larger $\varepsilon$ and $\delta$ will correspond to shorter preconditioner computation, we may conclude that, for a fixed $\alpha$, there will be an optimal combination $\varepsilon^\ast,\delta^\ast$ in the vicinity of the diagonal $\varepsilon=\delta$ that will minimise the overall runtime to solution. We further observe that no notable reductions in solver steps are achieved for parameter combinations $\varepsilon,\delta \ll \varepsilon^\ast\approx\delta^\ast$.

\begin{figure}[H]
    \centering
    \includegraphics[width=\columnwidth]{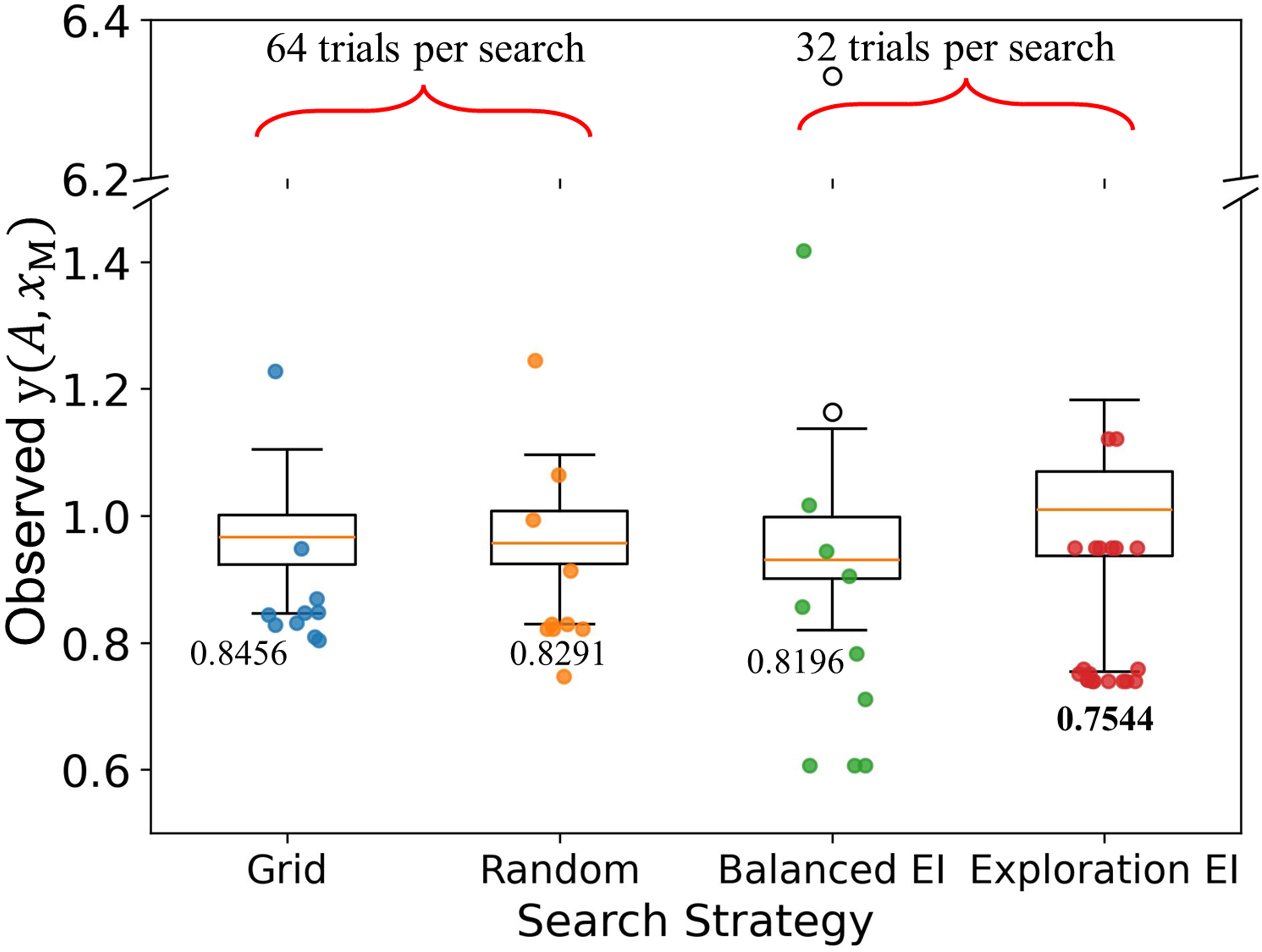}
    \caption{Box plot of sample median of y$(A,x_{\text{M}})$ over the explored $x_{\text{M}}$, including the minimum. Coloured circle points represent the distribution of the observed y$(A, x_{\text{M}}^*)$ over 10 replications, where $x_{\text{M}}^*$ indicates the parameter yielding the minimum sample median.}
    \Description{Box plot of sample median of y$(A,x_{\text{M}})$ over the explored $x_{\text{M}}$, including the minimum. Coloured circle points represent the distribution of the observed y$(A, x_{\text{M}}^*)$ over 10 replications, where $x_{\text{M}}^*$ indicates the parameter yielding the minimum sample median.}
    \label{fig:box_plots_fig}
\end{figure}

Finally, we assessed the practical utility of algorithmic parameter search for MCMC preconditioning on the ill-conditioned testing matrix \path{unsteady_adv_diff_order2_0001}. Despite using only 50\% of the evaluation budget (32 recommendations) compared to grid or random search (64 recommendations), the BO-enhanced recommendations reduced the number of steps to convergence through MCMC preconditioning by up to 25\%, which is approximately 10\% fewer steps than grid search (Figure~\ref{fig:box_plots_fig}). The box plot summarises the distribution of the sample medians from 10 replications over the explored candidates of algorithmic parameters $x_{\text{M}}$. In addition, the coloured circle points show the distribution of observations $y(A, x_{\text{M}}^*)$ over 10 replications, where $x_{\text{M}}^*$ denotes the single best recommendation of the algorithmic parameter that yields the minimum sample median of the MCMC preconditioning performance metric among all explored candidates for each search strategy.

\section{Conclusions}
We have introduced a BO framework that couples a graph neural surrogate with an EI-based acquisition strategy to tune MCMC preconditioners for solving linear systems. On an ill-conditioned matrix unseen during the training phase, the method required only 50\% of the search budget of a coarse grid yet reduced Krylov iterations by $\sim\!10\%$. These results demonstrate that structural information extracted from the matrix $A$ can guide hyper parameter search more efficiently than exhaustive sampling.

Several directions could further widen the scope and impact of the framework. First, although the Krylov method (GMRES, BiCGStab, CG) was included as a categorical input to the surrogate model, in this work we did not attempted to recommend the solver itself. Extending the framework to make such recommendations — selecting both the best solver and its optimal $(\alpha,\varepsilon,\delta)$ given inexpensive matrix features such as symmetry, approximate condition number, and sparsity — would be a next step. Also, practical deployments must balance iteration speed-up against preconditioner build time; this could be achieved by adding parameters that govern the filling fraction and truncation threshold, whose costs scale linearly with non-zeros and retained elements, respectively, as well as hardware knobs such as thread count and MPI ranks. Such considerations can be especially challenging in substructuring eigenvalue solvers due to the implicit nature of the Schur complement matrix \cite{kalantzis2020domain,kalantzis2021fast,kalantzis2023enhanced}. Also, future work could extend the framework to distributed-memory settings, explicitly minimising latency and accounting for communication and memory-management overheads to achieve robust scalability on parallel clusters.

Furthermore, the current FC layer-based forecasting in the last few layers of the graph neural surrogate model could be improved by replacing it with deep kernel learning \cite{wilson2016deep} or a scalable Gaussian Process layer \cite{liu2020gaussian} for forecasting the MCMC preconditioning metric and its uncertainty. This, in turn, would enhance the quality of EI. In addition, consideration of reinforcement learning approaches that propose a set of correlated MCMC algorithmic parameter vectors could exploit structure in the search space more effectively than independent EI maximisation. Furthermore, an active learning loop or generative model that generates new linear systems for evaluation would allow the surrogate to evolve continually towards broader classes of matrices. On the other hand, the practical utility of the framework could be improved by employing a cost-sensitive, weighted loss that places greater weight on larger systems, thereby yielding better parameter recommendations for them. Finally, identifying when to switch from full retraining to fine-tuning of the surrogate model is an important direction for improving the framework's practical applicability. Pursuing these directions will ultimately pave the way for more general and efficient linear-system solvers that accelerate scientific discovery across a broad range of applications such as climate modelling, computational fluid dynamics and plasma physics.

\begin{acks}
This work was supported by the Hartree National Centre for Digital Innovation, a UK Government-funded collaboration between STFC and IBM.
\end{acks}

\bibliographystyle{ACM-Reference-Format}
\bibliography{sample-base}


\end{document}